\ifdefined\pdfoutput\pdfoutput=1\fi

\documentclass[11pt]{article}

\usepackage[final]{acl}

\usepackage{times}
\usepackage{latexsym}
\usepackage{hyperref}
\usepackage{float}
\usepackage[T1]{fontenc}

\usepackage[utf8]{inputenc}

\usepackage{microtype}

\usepackage{inconsolata}

\usepackage{graphicx}

\title{Consolidating and Developing Benchmarking Datasets for the Nepali Natural Language Understanding Tasks}

\author{
  \textbf{*Jinu Nyachhyon\textsuperscript{1,2}},
  \textbf{*Mridul Sharma\textsuperscript{1,2}},
  \textbf{Prajwal Thapa\textsuperscript{1,2}},
  \textbf{Bal Krishna Bal\textsuperscript{1}}
\\
\\
  \textsuperscript{1}Information and Language Processing Research Lab (ILPRL), Kathmandu University
\\
  \textsuperscript{2}Institute for Research and Innovation in Intelligent Systems (IRIIS)
\\
  \small{
    \textbf{Correspondence:} \href{mailto:nyachhyonjinu@gmail.com}{nyachhyonjinu@gmail.com}, \href{mailto:mridulsharma3301@gmail.com}{mridulsharma3301@gmail.com}
  }
}

\begin{document}

\maketitle
\begin{abstract}
\begingroup
\renewcommand\thefootnote{}\footnote{* means equal contributions}%
\addtocounter{footnote}{-1}%
\endgroup
The Nepali language has distinct linguistic features, especially its complex script (Devanagari script), morphology, and various dialects, which pose a unique challenge for Natural Language Understanding (NLU) tasks. While the Nepali Language Understanding Evaluation (Nep-gLUE) benchmark provides a foundation for evaluating models, it remains limited in scope, covering four tasks. This restricts their utility for comprehensive assessments of Natural Language Processing (NLP) models. To address this limitation, we introduce twelve new datasets, creating a new benchmark, the Nepali Language Understanding Evaluation (NLUE) benchmark for evaluating the performance of models across a diverse set of Natural Language Understanding (NLU) tasks. The added tasks include Single-Sentence Classification, Similarity and Paraphrase Tasks, Natural Language Inference (NLI), and General Masked Evaluation Task (GMET). Through extensive experiments, we demonstrate that existing top models struggle with the added complexity of these tasks. We also find that the best multilingual model outperforms the best monolingual models across most tasks, highlighting the need for more robust solutions tailored to the Nepali language. This expanded benchmark sets a new standard for evaluating, comparing, and advancing models, contributing significantly to the broader goal of advancing NLP research for low-resource languages.
\end{abstract}

\section{Introduction}
\begin{table*}[t]
\centering
\resizebox{\textwidth}{!}{%
\begin{tabular}{llllll}
\hline
\textbf{Corpus} & \textbf{Train} & \textbf{Test} & \textbf{Task} & \textbf{Metrics Used} & \textbf{Domain} \\
\hline
\multicolumn{6}{c}{\textbf{Single Sentence Tasks}} \\
\hline
SA    & 65.1K  & 16.3K  & Sentiment                                   & Macro F1, Acc                      & Reviews, Tweets \\
CoLA  & 7.8K   & 1.95K  & Acceptability                               & Macro F1, Acc                      & Books, Journal \\
WG    & 32.5K  & 8.14K  & \begin{tabular}{@{}l@{}}Commonsense Reasoning \\ and Pronoun Coreference\end{tabular}
      & Macro F1, Acc                    & Misc. \\
\hline
\multicolumn{6}{c}{\textbf{Similarity and Paraphrase Tasks}} \\
\hline
QQP   & 26K    & 6.5K   & Paraphrase                                   & Macro F1, Acc                      & Social QA \\
MRPC  & 4.19K  & 1.05K  & Paraphrase                                   & Macro F1, Acc                      & News \\
STS-B  & 5.45K  & 1.36K  & Sentence Similarity                          & \begin{tabular}{@{}l@{}}Pearson Corr, \\ Spearman Corr, \\ $R^2$ \end{tabular} & News, Video Cap. \\
QADSM & 59.4K  & 14.9K  & Similarity                                   & Macro F1, Acc                      & News \\
\hline
\multicolumn{6}{c}{\textbf{Natural Language Inference Tasks}} \\
\hline
MNLI  & 40.8K  & 10.2K  & NLI                                          & Macro F1, Acc                      & Misc. \\
QNLI  & 28K    & 7K     & QA/NLI                                       & Macro F1, Acc                      & Wikipedia \\
RTE   & 2.01K  & 503    & NLI                                          & Macro F1, Acc                      & News, Wikipedia \\
CR    & 564    & 142    & Coreference/NLI                              & Macro F1, Acc                      & Fiction Books \\
\hline
\multicolumn{6}{c}{\textbf{General Masked Evaluation Task}} \\
\hline
GMET  & -      & 1.5K   & Mask Filling                                 & Acc, Combined Score
                    & Books, News \\
\end{tabular}
}
\caption{\label{tab:nlu-tasks}
Task descriptions and dataset statistics in the NLUE benchmark.
}
\end{table*}

Nepali is written in the Devanagari script and is a highly inflected language. The Nepali language incorporates a complex system of noun, adjective, and verb inflections, including gender, case, and number \cite{Bal2004}. It has a rich vocabulary and is spoken in different dialects across various regions, and there are variations in vocabulary, grammar, and pronunciation. Developing and establishing robust models for Nepali requires reliable methods to evaluate their quality and effectiveness and it is essential to have tools that can assess how well these models address the language's unique linguistic challenges while identifying their limitations.

Despite Nepali’s importance as a primary or secondary language for millions of speakers, research efforts and resources dedicated to its computational processing and evaluation remain relatively sparse. Existing benchmarks, such as Nep-gLUE \cite{timilsina-etal-2022-nepberta}, have made significant progress in this direction, providing a foundation for evaluating models on fundamental tasks. However, these benchmarks are limited in scope, primarily addressing four basic tasks and overlooking critical aspects of linguistic understanding such as coreference resolution, paraphrase interpretation, and advanced inference capabilities. To address this need, we introduce a new benchmark comprising 12 Natural Language Understanding (NLU) tasks \footnote{\href{https://huggingface.co/collections/IRIIS-RESEARCH/nepali-lanuguage-understanding-evaluation-benchmark-68592c0105fe37d5d97629d4}{NLUE Benchmark Datasets}} for Nepali. The tasks are grouped into four categories:
\begin{description}
  \item[Single-Sentence Tasks:] Sentiment Analysis (SA), Corpus of Linguistic Acceptability (CoLA), and WinoGrande (WG)
  \item[Similarity and Paraphrase Tasks:] Quora Question Pairs (QQP), Microsoft Research Paraphrase Corpus (MRPC), Semantic Textual Similarity Benchmark (STS-B), and Query-Ad Matching (QADSM)
  \item[Natural Language Inference (NLI) Tasks:] Multi-Genre NLI (MNLI), Question Answer NLI (QNLI), Recognizing Textual Entailment (RTE), and Coreference Resolution (CR)
  \item[General Masked Evaluation Task (GMET):] A diagnostic task for testing factual and contextual understanding.
\end{description}
This suite includes a broader range of linguistic tasks, enabling more comprehensive evaluation of NLU capabilities for Nepali language models (Appendix \ref{sec:appendixapoint1}). Table~\ref{tab:nlu-tasks} provides an overview of tasks, dataset sizes, evaluation metrics, and domains covered in the NLUE Benchmark.

The datasets in the NLUE Benchmark are inspired by the General Language Understanding Evaluation (GLUE) benchmark \cite{wang-etal-2018-glue} and XGLUE benchmark \cite{liang-etal-2020-xglue}, and were developed through a combination of automated and manual processes to ensure high-quality task-specific datasets. Our contributions involved translating datasets with Large Language Models (LLMs), particularly GPT-4o-mini \cite{4o-mini} and Gemini-2.5-flash \cite{2.5-flash}, and ensuring the accuracy and contextual relevance of these translations (Appendix \ref{sec:appendixapoint2} \& \ref{sec:appendixapoint3}). We also conducted a thorough review of the availability of existing Nepali datasets for each task. Where datasets were available, we integrated them with translated data, carefully eliminating duplicates to form a unified and comprehensive dataset. For tasks like Acceptability Judgments and Coreference Resolution, where suitable datasets or high-quality translations were unavailable, we performed manual translations to ensure linguistic accuracy and consistency. These efforts collectively ensure that the final dataset is robust, comprehensive, and reflective of the linguistic diversity in the Nepali language.

To assess the effectiveness of the NLUE benchmark and performance of models, we conducted experiments by fine-tuning both monolingual models trained exclusively on Nepali-language data and multilingual models that include Nepali as one of their supported languages. Each model was fine-tuned on tasks introduced in the NLUE Benchmark and evaluated using metrics provided in \autoref{tab:nlu-tasks}, providing a comprehensive understanding of their performance on various aspects of NLU.

\section{Related Works}

Benchmarks like GLUE \cite{wang-etal-2018-glue} and its successor Super General Language Understanding Evaluation (SuperGLUE) benchmark \cite{wang2020superglue} have been instrumental in advancing research in Natural Language Understanding (NLU). GLUE introduced a multi-task framework for evaluating diverse NLU capabilities, such as single-sentence classification, sentence-pair similarity, and inference tasks. SuperGLUE extended this with more challenging tasks, including causal reasoning and coreference resolution, addressing GLUE's limitations for state-of-the-art models. These benchmarks set a standard for evaluating linguistic and semantic understanding in high-resource languages like English, inspiring adaptations in other languages and low-resource settings. Efforts like \cite{liang-etal-2020-xglue} and \cite{hu2020xtreme} expanded these concepts to multilingual contexts, enabling cross-lingual transfer learning.

Nep-gLUE \cite{timilsina-etal-2022-nepberta} is the first comprehensive benchmark for Natural Language Understanding (NLU) tasks in Nepali. It includes four core tasks: Named Entity Recognition (NER), Part-of-Speech Tagging (POS), Content Classification (CC), and Categorical Pair Similarity (CPS). Although Nep-gLUE offers a robust foundation with its multi-task dataset, it falls short in addressing more advanced NLP tasks necessary for comprehensive evaluations of models at the linguistic level. The advanced and complex tasks are crucial for further progress in low-resource languages like Nepali.

Nepali Sentiment Analysis (NepSA) \cite{singh-2020} is a targeted aspect-based sentiment analysis dataset, with 3,068 comments extracted from 37 YouTube videos across 9 channels. It is annotated using a binary sentiment polarity schema across six aspects: General, Profanity, Violence, Feedback, Sarcasm, and Out-of-scope. Another dataset, Aspect-Based Sentiment Analysis \cite{tamrakar-2020}, contains 1,576 sentences, equally divided between positive and negative sentiments. Additional datasets, such as Nepali Language Sentiment Analysis - Movie Reviews \cite{Shikharkaggle} with 602 data points, and Nepali Sentiment Analysis \cite{Maheshkaggle} with 2,161 data points found on Kaggle, are limited in size and domain-specific. For our benchmark, we utilized the NepCOV19Tweets dataset \cite{sitaula-2021}, which includes \textasciitilde 33.5k sentiments labeled as positive, negative, or neutral. From these, we selected 14.9k positive and 13.5k negative data points for the SA dataset. A more recent dataset, Sentiment of Election-Based Nepali Tweets \cite{Durgakaggle}, contains \textasciitilde 17.8k tweets but includes English characters and numbers, making it less suitable for our benchmarked dataset. To our knowledge, there are no publicly available datasets for coreference resolution, acceptability judgment, paraphrase and similarity detection, commonsense reasoning, pronoun coreference resolution, general masked evaluation, or NLI in the Nepali language. Despite some studies focusing on Nepali grammar, the lack of datasets for these advanced tasks limits the development of comprehensive NLU benchmarks.

\section{Model Selection}

To evaluate the performance of Natural Language Processing (NLP) models on the Nepali Language Understanding Evaluation (NLUE) benchmark, we selected ten publicly available models that support devanagari script, carefully chosen to represent a diverse range of architectures, parameter sizes, and pretraining strategies including the state-of-the-art encoder model for language understanding and best monolingual models for the Nepali Language. Evaluating these models on the NLUE benchmark serves multiple purposes. First, it provides a comprehensive assessment of their capabilities across a diverse set of tasks. This enables us to identify which architectures and pretraining strategies are best suited for Nepali NLP, particularly for tasks that demand robust handling of the language’s morphological complexity and dialectal variations. Second, comparing monolingual and multilingual models highlights the trade-offs between language-specific pretraining and cross-lingual generalization, offering insights into the optimal approach for low-resource languages. By identifying the strengths and weaknesses of existing models, this study informs the development of more robust solutions tailored to Nepali’s unique linguistic challenges.

\section{Tasks}
NLUE is a benchmark designed to evaluate the performance of language understanding models across a diverse set of tasks, addressing the limitations of its predecessor, Nep-gLUE. The objective of NLUE is to provide a robust evaluation metric applicable to a broad range of language understanding challenges. We describe the tasks below and in \autoref{tab:nlu-tasks}.

\subsection{Single-Sentence Tasks}
\begin{table*}[t]
\centering
\resizebox{\textwidth}{!}{%
\begin{tabular}{lcccccccc}
\hline
\textbf{Model} & \textbf{Params} & \multicolumn{2}{c}{\textbf{SA}} & \multicolumn{2}{c}{\textbf{CoLA}} & \multicolumn{2}{c}{\textbf{WG}} \\
              &                 & Acc   & F1    & Acc   & F1    & Acc   & F1 \\
\hline
Distilbert-Nepali~\cite{maskey-etal-2022-nepali}   & 67M  & 86.34 & 86.33 & 84.51 & 80.96 & 58.20 & 58.08 \\
NepBERT~\cite{rajan2021github}             & 82M  & 83.34 & 83.34 & 80.51 & 74.80 & 52.49 & 52.04 \\
NepaliBERT~\cite{Pudasaini2023}          & 110M & 87.06 & 87.06 & 84.51 & 80.92 & 54.77 & 54.75 \\
BERT Nepali~\cite{thapa-etal-2025-development}         & 110M & 87.73 & 87.72 & 84.76 & 80.65 & 66.81 & 66.13 \\
NepBERTa~\cite{timilsina-etal-2022-nepberta}             & 110M & 86.62 & 86.62 & 84.15 & 80.60 & 67.12 & 50.52 \\
RoBERTa Nepali~\cite{thapa-etal-2025-development}      & 125M & 87.75 & 87.74 & 85.44 & 82.14 & \textbf{68.07} & \textbf{68.07} \\
DeBERTa-Nepali~\cite{maskey-etal-2022-nepali}      & 139M & 87.43 & 87.42 & 85.08 & 81.86 & 59.76 & 59.75 \\
Multilingual BERT~\cite{Devlin2019}   & 172M & 86.35 & 86.34 & 82.41 & 78.95 & 67.12 & 50.52 \\
XLM-R Base~\cite{conneau-etal-2020-unsupervised}          & 270M & 88.33 & 88.34 & 85.64 & 82.03 & 50.77 & 50.52 \\
m-DeBERTa-v3~\cite{he2023debertav3improvingdebertausing}        & 276M & \textbf{88.94} & \textbf{88.93} & \textbf{88.31} & \textbf{85.64} & 67.45 & 67.44 \\
\hline
\end{tabular}
}
\caption{\label{tab:results-sa-cola-wg}
Model Performance across Single-Sentence Tasks
}
\end{table*}

Single-sentence tasks in the NLUE benchmark focus on assessing a model's ability to understand and analyze individual sentences. These tasks evaluate a model's ability to understand and interpret the meaning, sentiment, and grammatical structure of individual sentences.

\subsubsection{Sentiment Analysis (SA)}
The Sentiment Analysis dataset has been added to evaluate models' ability to classify the emotional tone (Positive \& Negative) of Nepali text. We created the dataset for sentiment analysis by translating Stanford Sentiment Treebank \cite{socher-etal-2013-recursive} from the GLUE Benchmark, which includes \textasciitilde 53k sentence-level data points from movie reviews with human-annotated sentiment labels, using GPT-4o-mini, and manually translating instances that could not be accurately translated (Appendix \ref{sec:appendixapoint2} \& \ref{sec:appendixapoint3}). We incorporated this dataset with pre-existing sentiment analysis of Nepali COVID-19-related tweets \cite{sitaula-2021}, adding \textasciitilde 28.4k data points. The combined SA dataset totals 81.4k data points, equally distributed between the positive and negative classes. Models are evaluated using Accuracy and Macro F1-score metrics, as reported in \autoref{tab:nlu-tasks}.

\subsubsection{Corpus of Linguistic Acceptability (CoLA)}
\begin{table*}[t]
\centering
\resizebox{\textwidth}{!}{%
\begin{tabular}{lcccccccccc}
\hline
\textbf{Model} & \textbf{Params} 
& \multicolumn{2}{c}{\textbf{QQP}} 
& \multicolumn{2}{c}{\textbf{MRPC}} 
& \multicolumn{3}{c}{\textbf{STS-B}} 
& \multicolumn{2}{c}{\textbf{QADSM}} \\
& & Acc & F1 & Acc & F1 & Sp. corr & Pr. corr & $R^2$ & Acc & F1 \\
\hline
Distilbert-Nepali~\cite{maskey-etal-2022-nepali}   & 67M  & 81.63 & 81.07 & 80.52 & 77.95 & 84.57 & 83.13 & 71.13 & 65.74 & 65.63 \\
NepBERT~\cite{rajan2021github}             & 82M  & 71.17 & 69.61 & 67.34 & 56.91 & 41.44 & 41.06 & 12.71 & 61.03 & 60.80 \\
NepaliBERT~\cite{Pudasaini2023}          & 110M & 77.37 & 76.67 & 66.86 & 58.49 & 75.91 & 73.76 & 57.62 & 63.27 & 63.18 \\
BERT Nepali~\cite{thapa-etal-2025-development}         & 110M & 80.88 & 80.48 & 76.50 & 73.61 & 84.57 & 83.87 & 71.54 & 64.35 & 64.34 \\
NepBERTa~\cite{timilsina-etal-2022-nepberta}             & 110M & 81.83 & 81.57 & 80.42 & 78.67 & 87.54 & 86.44 & 76.53 & 63.71 & 63.64 \\
RoBERTa Nepali~\cite{thapa-etal-2025-development}      & 125M & 81.15 & 80.60 & 79.47 & 75.25 & 87.75 & 86.32 & 76.68 & 65.02 & 65.00 \\
DeBERTa-Nepali~\cite{maskey-etal-2022-nepali}      & 139M & 82.85 & 82.33 & 80.61 & 78.23 & 81.62 & 80.00 & 66.35 & 65.23 & 65.21 \\
Multilingual BERT~\cite{Devlin2019}   & 172M & 82.31 & 81.78 & 81.47 & 78.02 & 87.75 & 86.62 & 76.93 & 63.91 & 63.84 \\
XLM-R Base~\cite{conneau-etal-2020-unsupervised}          & 270M & 83.06 & 82.68 & 82.71 & 80.59 & 87.68 & 86.79 & 76.77 & 63.70 & 63.70 \\
m-DeBERTa-v3~\cite{he2023debertav3improvingdebertausing}        & 276M & \textbf{84.34} & \textbf{83.82} & \textbf{83.48} & \textbf{81.93} & \textbf{90.22} & \textbf{89.57} & \textbf{81.33} & \textbf{66.42} & \textbf{66.42} \\
\hline
\end{tabular}
}
\caption{\label{tab:results-sim-paraphrase}
Model Performance across Similarity and Paraphrase Tasks 
}
\end{table*}

The Acceptability Judgments dataset determines whether a given sentence follows the linguistic rules of Nepali, ensuring the model can assess grammaticality. The dataset was created by translating the Corpus of Linguistic Acceptability (CoLA) \cite{warstadt-etal-2019-neural} from the GLUE Benchmark, which includes 9.75k data points sourced from books and journal articles on linguistic theory, using GPT-4o-mini. Manual corrections were applied to sections where translations were inaccurate (Appendix \ref{sec:appendixapoint2} \& \ref{sec:appendixapoint3}). The dataset is divided into correct and incorrect classes in a 70:30 ratio, respectively. Models are evaluated using Accuracy and Macro F1-score metrics, as reported in \autoref{tab:nlu-tasks}.

\subsubsection{WinoGrande (WG)}
The WinoGrande dataset evaluates a model’s ability to perform commonsense reasoning by identifying the correct referent in a sentence with a blank referring to one of two candidate entities. The dataset for this benchmark is converted to Nepali by translating the XGLUE benchmark’s WinoGrande dataset \cite{sakaguchi2019} using Gemini-2.5-flash, with manual corrections applied to ensure translation accuracy (Appendix \ref{sec:appendixapoint2} \& \ref{sec:appendixapoint3}). The final dataset contains 40.7k data points, with each instance labeled to indicate the correct referent, and is equally split between both classes. The dataset preserves the original format and balance of the English version. Models are evaluated using Accuracy and Macro F1-score metrics, as reported in \autoref{tab:nlu-tasks}.

\subsection{Similarity and Paraphrase Tasks}
Similarity and Paraphrase Task in the NLUE benchmark evaluates a model's ability to determine whether two sentences convey the same meaning or are paraphrases of each other. By focusing on this aspect of language comprehension, these tasks provide valuable insights into a model's proficiency in handling diverse expressions of similar ideas.

\subsubsection{Quora Question Pairs (QQP)}
The QQP dataset tests whether the model can identify if pairs of questions from the community question-and-answer website Quora have similar meanings. The dataset was created by translating the Quora Question Pairs dataset \cite{Quora2017} from the GLUE Benchmark into Nepali using GPT-4o-mini and Gemini-2.5-flash, with manual corrections applied (Appendix \ref{sec:appendixapoint2} \& \ref{sec:appendixapoint3}). The dataset contains 32.5k question pairs, labeled as similar or dissimilar, with a class distribution of 40\% similar and 60\% dissimilar. Models are assessed using accuracy and Macro F1-score metrics, as reported in \autoref{tab:nlu-tasks}.

\subsubsection{Microsoft Paraphrase Research Corpus (MPRC)}
We introduced the MRPC dataset, intending to identify whether the sentence pairs extracted from news articles are paraphrases of each other, based on the Microsoft Research Paraphrase Corpus \cite{dolan-brockett-2005-automatically}. Using GPT-4o-mini, we translated the MRPC dataset into Nepali with manual correction whenever needed (Appendix \ref{sec:appendixapoint2} \& \ref{sec:appendixapoint3}). The dataset contains 5.23k sentence pairs, with the class distribution of 70-30, with a higher proportion of paraphrase pairs. We report the Accuracy and Macro F1 score, as shown in \autoref{tab:nlu-tasks}.

\subsubsection{Semantic Textual Similarity Benchmark (STS-B)}
The STS-B dataset measures a model’s proficiency in predicting the degree of semantic relatedness between pairs of sentences drawn from sources such as news headlines and video captions. Each pair is annotated with a similarity score on a continuous scale from 0 (no meaning overlap) to 5 (complete semantic equivalence), framing the task as a regression problem. The dataset was created by translating the STS-B dataset \cite{cer-etal-2017-semeval} from the GLUE Benchmark into Nepali using Gemini-2.5-flash, with manual corrections applied to ensure translation accuracy (Appendix \ref{sec:appendixapoint2} \& \ref{sec:appendixapoint3}). The dataset contains 6.82k sentence pairs. We evaluate the model using Pearson correlation, Spearman correlation, and R\textsuperscript{2} metrics, as reported in \autoref{tab:nlu-tasks}.

\subsubsection{Query-Ad Matching (QADSM)}
The QADSM dataset assesses a model’s capability to align the semantic meaning between queries and advertisements. The dataset was created by translating the QADSM dataset from the XGLUE Benchmark \cite{liang-etal-2020-xglue} into Nepali using Gemini-2.5-flash, with manual refinements to ensure linguistic precision (Appendix \ref{sec:appendixapoint2} \& \ref{sec:appendixapoint3}). The dataset contains 74.3k data points, equally split between relevant and irrelevant classes, based on ad-query relevance. Models are evaluated using accuracy and Macro F1-score metrics, as reported in \autoref{tab:nlu-tasks}.

\subsection{Inference Tasks}
\begin{table*}[t]
\centering
\resizebox{\textwidth}{!}{%
\begin{tabular}{lcccccccccc}
\hline
\textbf{Model} & \textbf{Params} 
& \multicolumn{2}{c}{\textbf{MNLI}} 
& \multicolumn{2}{c}{\textbf{QNLI}} 
& \multicolumn{2}{c}{\textbf{RTE}} 
& \multicolumn{2}{c}{\textbf{CR}} \\
& & Acc & F1 & Acc & F1 & Acc & F1 & Acc & F1 \\
\hline
Distilbert-Nepali~\cite{maskey-etal-2022-nepali}   & 67M  & 68.57 & 68.61 & 79.61 & 79.46 & 56.06 & 55.63 & 52.82 & 52.63 \\
NepBERT~\cite{rajan2021github}             & 82M  & 51.93 & 51.09 & 60.04 & 59.96 & 53.88 & 53.85 & 55.63 & 39.70 \\
NepaliBERT~\cite{Pudasaini2023}          & 110M & 63.27 & 63.25 & 77.58 & 77.43 & 51.89 & 51.80 & 55.63 & 55.63 \\
BERT Nepali~\cite{thapa-etal-2025-development}         & 110M & 71.80 & 71.92 & 81.26 & 81.22 & 53.28 & 53.27 & 59.15 & 51.63 \\
NepBERTa~\cite{timilsina-etal-2022-nepberta}             & 110M & 71.87 & 71.85 & 81.24 & 81.15 & 55.07 & 53.33 & \textbf{58.52} & \textbf{57.33} \\
RoBERTa Nepali~\cite{thapa-etal-2025-development}      & 125M & 73.10 & 73.07 & 81.86 & 81.78 & 52.49 & 52.44 & 49.29 & 49.12 \\
DeBERTa-Nepali~\cite{maskey-etal-2022-nepali}      & 139M & 74.01 & 74.01 & 82.64 & 82.64 & 53.88 & 53.04 & 50.70 & 50.67 \\
Multilingual BERT~\cite{Devlin2019}   & 172M & 71.60 & 71.85 & 83.47 & 83.46 & \textbf{68.19} & \textbf{68.00} & 47.89 & 47.38 \\
XLM-R Base~\cite{conneau-etal-2020-unsupervised}          & 270M & 75.23 & 75.22 & 83.13 & 83.13 & 57.06 & 54.86 & 50.00 & 49.80 \\
m-DeBERTa-v3~\cite{he2023debertav3improvingdebertausing}        & 276M & \textbf{78.76} & \textbf{78.84} & \textbf{86.65} & \textbf{86.65} & 57.85 & 57.80 & 46.48 & 32.84 \\
\hline
\end{tabular}
}
\caption{\label{tab:results-nli}
Model Performance across Inference Tasks
}
\end{table*}

The NLI tasks in this benchmark assess a model's ability to understand relationships between sentences, such as entailment, contradiction, and neutral alignment. These tasks are crucial because they evaluate a model's comprehension of contextual meaning, logical inference, and its ability to handle complex linguistic structures, making them essential for advancing robust language understanding.

\subsubsection{Multi-Genre NLI (MNLI)}
The MNLI dataset tests a model’s capability to predict the relationship between sentence pairs, determining whether a premise entails, contradicts, or is unrelated to a hypothesis (neutral). The dataset was created by translating the Stanford Natural Language Inference Corpus \cite{bowman2015largeannotatedcorpuslearning} from the GLUE Benchmark into Nepali using GPT-4o-mini and Gemini-2.5-flash, with manual intervention for precision (Appendix \ref{sec:appendixapoint2} \& \ref{sec:appendixapoint3}). The dataset contains 51k sentence pairs, equally divided among entailment, contradiction, and neutral classes. We report accuracy and Macro F1-score, as described in \autoref{tab:nlu-tasks}.

\subsubsection{Question-Answering NLI (QNLI)} 
The QNLI dataset evaluates a model’s capability to determine whether a context sentence contains the answer to a given question. The dataset has been adapted for Nepali from the GLUE benchmark by translating the original English dataset using GPT-4o-mini and Gemini-2.5-flash, with manual verification for accuracy (Appendix \ref{sec:appendixapoint2} \& \ref{sec:appendixapoint3}), which originates from the Stanford Question Answering Dataset \cite{rajpurkar-etal-2016-squad} that contains question-paragraph pairs sourced from Wikipedia. The dataset contains 35k question-sentence pairs, equally split between entailment and non-entailment pairs, ensuring a balanced class distribution, and evaluated using accuracy and Macro F1-score metrics, as reported in \autoref{tab:nlu-tasks}.

\subsubsection{Recognizing Textual Entailment (RTE)}
The RTE dataset evaluates a model's ability to predict whether a hypothesis logically follows from a given premise. The dataset for this benchmark is converted to Nepali by translating the GLUE benchmark’s RTE dataset, combined from RTE1 \cite{Dagan2006}, RTE2 \cite{BarHaim2006TheSP}, RTE3 \cite{Giampiccolo2007TheTP}, and RTE5 \cite{bentivogli2009fifth} using GPT-4o-mini, with manual corrections to maintain translation accuracy (Appendix \ref{sec:appendixapoint2} \& \ref{sec:appendixapoint3}), containing 2.51k data points, equally distributed between two classes (entailment and non-entailment). We evaluate the model using Accuracy and Macro F1-score, as discussed in \autoref{tab:nlu-tasks}.

\subsubsection{Coreference Resolution (CR)} 
This CR dataset tests the model’s ability to resolve coreference relationships within a Nepali text. We developed the coreference resolution dataset by manually translating the Winograd Schema Challenge \cite{Levesque2011TheWS} from the GLUE Benchmark. The dataset has 706 data points, balanced between two classes, evaluated with Accuracy and Macro F1-score, as mentioned in \autoref{tab:nlu-tasks}.

\subsection{General Masked Evaluation Task (GMET)}

The General Masked Evaluation Task (GMET) dataset evaluates the zero-shot capabilities of language models in recognizing word relationships, understanding contextual nuances, and maintaining grammatical precision without fine-tuning. It serves as a benchmark for assessing logical reasoning and proficiency with complex linguistic constructs, critical for reliable language understanding across diverse scenarios. The GMET dataset comprises 1,500 authentic sentences from real-world contexts, ensuring ecological validity. These sentences are organized into 75 distinct categories, with 20 sentences per category, covering various topics and regional linguistic variations. Each sentence contains a missing word, challenging models to predict the appropriate word based on context, testing their inherent contextual understanding and language comprehension, particularly with nuanced expressions across communities. As the missing word may not always have a single correct answer, native speakers assisted in manual evaluations to ensure accurate and fair assessment.

\begin{table}[H]
\centering
\small
\setlength{\tabcolsep}{1pt}
\resizebox{\columnwidth}{!}{ 
\begin{tabular}{l@{\hspace{1pt}}c@{\hspace{1pt}}c@{\hspace{1pt}}c}
\hline
\textbf{Model} & \textbf{Params} & \textbf{Acc} & \textbf{C. Acc} \\
\hline
Distilbert-Nepali~\cite{maskey-etal-2022-nepali} & 67M & 51.47 & 42.84 \\
NepBERT~\cite{rajan2021github} & 82M & 13.60 & 12.52 \\
NepaliBERT~\cite{Pudasaini2023} & 110M & 44.53 & 37.99 \\
BERT Nepali~\cite{thapa-etal-2025-development} & 110M & 49.40 & 42.63 \\
NepBERTa~\cite{timilsina-etal-2022-nepberta} & 110M & 46.40 & 39.89 \\
RoBERTa Nepali~\cite{thapa-etal-2025-development} & 125M & \textbf{57.27} & \textbf{48.76} \\
DeBERTa-Nepali~\cite{maskey-etal-2022-nepali} & 139M & 52.60 & 44.56 \\
Multilingual BERT~\cite{Devlin2019} & 172M & 14.13 & 12.80 \\
XLM-R Base~\cite{conneau-etal-2020-unsupervised} & 270M & 53.27 & 44.75 \\
m-DeBERTa-v3~\cite{he2023debertav3improvingdebertausing} & 276M & 45.33 & 42.77 \\
\hline
\end{tabular}
}
\caption{\label{tab:gmet-results}Model Performance in GMET}
\end{table}

Model performance on the GMET dataset is evaluated using two key metrics: overall accuracy and a combined score. Overall accuracy measures the proportion of correct predictions across all sentences, providing a straightforward performance indicator. The combined score integrates overall accuracy with an equality score, reflecting consistency across categories and penalizing uneven performance to ensure balanced contextual understanding across diverse topics and linguistic variations. Further details are provided in Appendix \ref{sec:appendixapoint6}.

\section{Experiments}
\label{sec:experiments}

\begin{figure}[H]
  \includegraphics[width=\columnwidth]{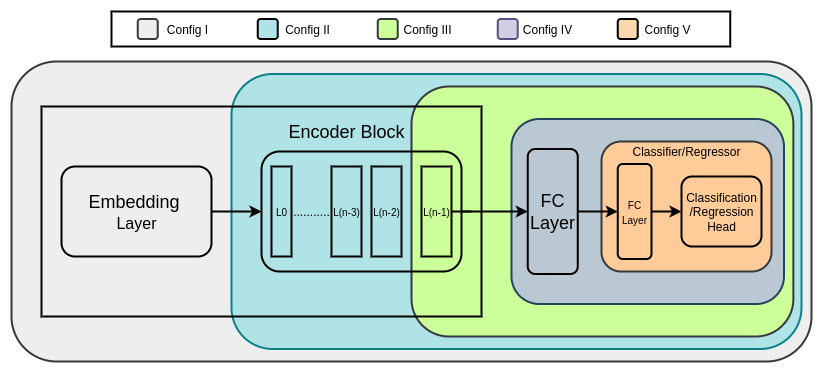}
  \caption{Different training config based on parameters with initial FC Layer}
  \label{fig:training_config_with_FC}
\end{figure}

\begin{figure}[H]
  \includegraphics[width=\columnwidth]{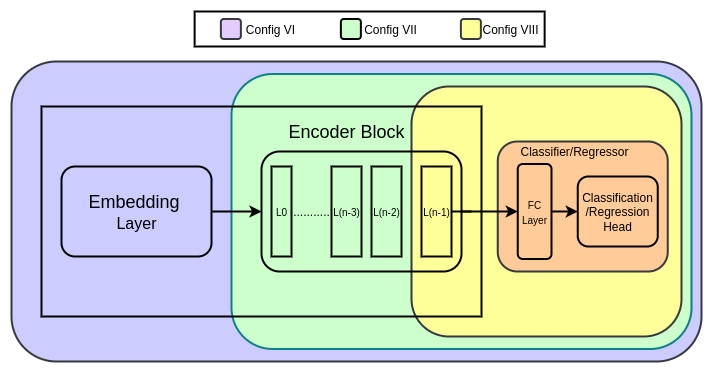}
  \caption{Different training config based on parameters without initial FC Layer}
  \label{fig:training_config_without_FC}
\end{figure}

We experimented with eight distinct finetuning configurations (Configs I–VIII), each controlling the subset of model parameters that are updated during training, as illustrated in \autoref{fig:training_config_with_FC} and \autoref{fig:training_config_without_FC}. These configuration choices were driven by our available datasets to mitigate overfitting risks, which ranged widely from just under 1,000 to 80,000 data points. For larger datasets like QADSM and SA, training only the classification layer was insufficient, while for smaller datasets like CR and RTE, training all layers risked overfitting. Therefore, at least three configurations were tested per dataset to ensure robust performance comparisons and validate generalizability.

As part of our ablation study, we systematically examined how performance was affected by varying the number and type of layers updated during training, ranging from tuning only the top classification layer to progressively unfreezing intermediate and lower transformer layers. This analysis helped isolate the contributions of different layers to downstream performance. Specifically, we also experimented with Config IV both with and without the initial fully connected (FC) layer to assess its specific role in feature transformation.

Hyperparameter Search Space:
\begin{itemize}
    \item Learning rate: $\{1e-5, 2e-5, 1e-4, 2e-4\}$
    \item Batch size: $\{8, 16, 32\}$
    \item Training epochs: Up to 15, with early stopping after three consecutive epochs of non-improving validation loss
\end{itemize}

For each configuration, we performed 5-fold cross-validation to select optimal hyperparameters and evaluate model performance. For each fold, we trained models with all hyperparameter combinations and selected the configuration that achieved the lowest validation loss. The best average hyperparameters across folds were used for final training. Test evaluation was done only after hyperparameter selection. Optimal hyperparameter settings and configuration for each dataset and model are reported in Appendix \ref{sec:appendixapoint5}.

\section{Result and Analysis}
We evaluate 10 language models on the NLUE benchmark across four task categories: Single-Sentence Classification, Similarity and Paraphrase Detection, Inference Tasks, and the GMET. For Classification Tasks, we report accuracy to measure overall correctness and macro-F1 to ensure balanced performance across potentially imbalanced classes. For the Regression Task, we use Spearman and Pearson correlation coefficients to assess monotonic and linear relationships, respectively, between predicted and actual continuous scores, and $R^2$ to quantify the proportion of variance in actual similarity scores explained by the model’s predictions. For GMET, we report a combined score, integrating overall accuracy with an equality score, to evaluate consistency across diverse categories.

Across Single Sentence tasks, m-DeBERTa-v3 achieves the highest overall scores, with an SA Macro-F1 of 88.93, WG Macro-F1 of 67.44, and CoLA Macro-F1 of 85.64. Among Nepali-specific models, RoBERTa-Nepali performs competitively in SA and WG, indicating that moderate-scale models can effectively handle single-sentence understanding in Nepali. Results reported in \autoref{tab:results-sa-cola-wg}.

m-DeBERTa-v3 consistently achieves the highest scores across Similarity and Paraphrase Tasks with top Macro-F1 scores in QQP (83.82), MPRC (81.93), QADSM (66.42), and the highest correlation metrics in STS-B (90.22 Spearman, 89.57 Pearson). Among Nepali-specific models, DeBERTa-Nepali performs strongly on QQP and MRPC, while RoBERTa-Nepali shows better results on STS-B and QADSM. Overall, multilingual models dominate in performance, which suggests that semantic similarity detection in Nepali demands sophisticated representational capabilities beyond what current Nepali-specific models provide. Results reported in \autoref{tab:results-sim-paraphrase}.

In Inference Tasks, m-DeBERTa-v3 achieves the strongest performance on MNLI (78.84 Macro-F1) and QNLI (86.65 Macro-F1), while Multilingual BERT achieves a strong 68 Macro-F1 on RTE, suggesting that multilingual pretraining enhances entailment and contradiction processing capabilities. However, their performance drops notably on the CR task, with no model exceeding 59.15\% Accuracy (BERT Nepali), mainly due to its complexity in Nepali and limited dataset size (706 data points), which indicates that all models struggle with generalization from small datasets. Results reported in \autoref{tab:results-nli}.

All evaluated models demonstrated suboptimal performance on the GMET dataset. These results indicate that zero-shot tasks in Nepali present significant challenges for current language models, even when processing straightforward conversational sentences. Notably, multilingual models and those with larger parameter counts failed to achieve superior performance compared to their monolingual counterparts. This performance gap may be attributed to tokenization limitations, as multilingual models typically contain fewer Devanagari tokens in their vocabularies relative to monolingual Nepali models. Results reported in \autoref{tab:gmet-results}.

\section{Conclusion}
The NLUE benchmark reveals distinct performance trends across models and tasks, with model size correlating strongly with performance. Larger models with multilingual pretraining, such as m-deberta-v3 (276M parameters) and XLM-r-base (270M parameters), consistently outperform smaller Nepali-specific models, particularly in tasks requiring nuanced semantic understanding (e.g., STS-B, QNLI). However, RoBERTa-Nepali (125M parameters) achieves competitive results despite its smaller size, suggesting that quality pretraining can outweigh parameter count.

Tasks like RTE and CR remain challenging, due to smaller dataset sizes, highlighting the need for enhanced datasets and improved modeling of Nepali textual entailment and coreference resolution. These results underscore the potential of multilingual models for low-resource languages like Nepali, while also revealing the importance of better Nepali-specific models to address language-specific challenges. Future work should prioritize creating larger, more diverse Nepali datasets and exploring techniques like cross-lingual transfer to enhance model robustness. The NLUE benchmark provides a valuable framework for evaluating and improving language models, paving the way for advancements in Nepali NLP.

\section{Limitations}
While the Nepali Language Understanding Evaluation (NLUE) benchmark significantly advances the evaluation of Natural Language Processing models for the Nepali Language, several limitations must be acknowledged to contextualize the findings and guide future research.

First, the datasets introduced in the NLUE benchmark were primarily created by translating existing English-language datasets from benchmarks such as GLUE and XGLUE, using automated tools like GPT-4o-mini and Gemini2.5-flash, supplemented by manual corrections. Although efforts were made to ensure translation accuracy, subtle linguistic nuances, cultural contexts, and idiomatic expressions specific to Nepali may not have been fully captured. The small size of certain datasets (e.g., CR and RTE) limits model performance and shows models' lack of generalization on smaller datasets. Second, the study evaluates a range of models with varying parameter sizes (67M to 276M), but resource constraints prevented the inclusion of larger, state-of-the-art models or extensive hyperparameter tuning. Finally, the reliance on specific evaluation metrics (e.g., accuracy, Macro-F1 score, Spearman, and Pearson correlations) may not fully capture the models’ performance across all dimensions of language understanding. For example, the GMET task relies on manual evaluations by native speakers, which might introduce subjectivity and potential inconsistencies.

\bibliography{custom}

\appendix

\section{Dataset Description}
\label{sec:appendixapoint1}

\subsection{Sentiment Analysis (SA)}
To evaluate the sentiment understanding capabilities of language models, we developed a sentiment analysis dataset by combining existing Nepali sentiment datasets and translating several high-quality examples from English to Nepali.

\begin{figure}[H]
  \centering
  \includegraphics[width=1\columnwidth]{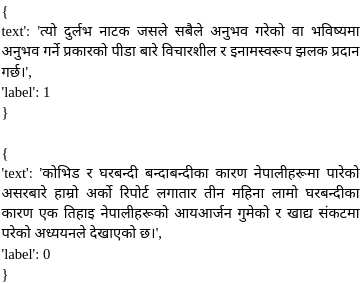}
  \caption{SA Positive (1) and Negative (0) Sample}
  \label{fig:SA_exp}
\end{figure}

Sentiment analysis requires models to grasp not just the literal meaning of words but also their emotional undertones and contextual implications. It is particularly challenging in Nepali, where sentiment is often conveyed through subtle linguistic cues and cultural context that may not be immediately apparent. Using this, we can better understand whether models truly comprehend the nuanced ways emotions are expressed in Nepali text, rather than simply memorizing surface-level patterns. This evaluation helps us gauge how well these models might perform on real-world applications involving subjective content analysis.

\subsection{Corpus of Linguistic Acceptability (CoLA)}
The Corpus of Linguistic Acceptability (CoLA) is a dataset originally developed for the GLUE benchmark to assess a model’s ability to judge the grammatical acceptability of English sentences. We incorporated CoLA-Nepali into our evaluation suite because understanding grammatical structure is fundamental to language comprehension. Unlike other benchmarks that primarily test semantic understanding or task performance, CoLA directly probes whether models have internalized the syntactic rules that govern sentence formation.

\begin{figure}[H]
  \centering
  \includegraphics[width=1\columnwidth]{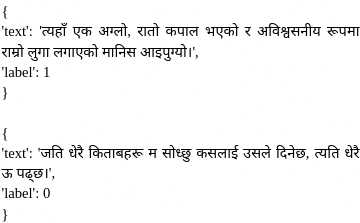}
  \caption{CoLA Positive (1) and Negative (0) Sample}
  \label{fig:COLA_exp}
\end{figure}

This evaluation is especially crucial for Nepali, where limited training data often contains grammatical inconsistencies or errors. By testing linguistic acceptability, we can determine whether our models have learned to distinguish well-formed Nepali sentences from those that violate grammatical constraints. This provides valuable insight into how deeply the models understand Nepali's structural patterns, beyond their ability to perform specific NLP tasks.

\subsection{WinoGrande (WG)}
The WinoGrande dataset is a large-scale collection of Winograd Schema Challenge-style problems designed to evaluate commonsense reasoning and contextual disambiguation in natural language understanding. 

\begin{figure}[H]
  \centering
  \includegraphics[width=1\columnwidth]{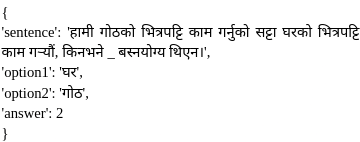}
  \caption{WG Sample}
  \label{fig:WG_exp}
\end{figure}

We translated \& added the WinoGrande dataset into our evaluation benchmark to assess models' ability to perform commonsense reasoning and resolve linguistic ambiguities. The dataset’s adversarial construction reduces reliance on superficial statistical patterns, ensuring models rely on deep semantic and commonsense reasoning, which is essential for real-world applications like dialogue systems or conversational agents in Nepali. Including WinoGrande in the Nepali benchmark allows us to evaluate model strengths and limitations in handling nuanced linguistic structures and cultural contexts specific to Nepali, thereby improving their robustness for practical, context-sensitive applications.

\subsection{Quora Question Pairs (QQP)}
This dataset consists of pairs of questions that are labeled as either paraphrases (semantically equivalent) or not.

\begin{figure}[H]
  \centering
  \includegraphics[width=1\columnwidth]{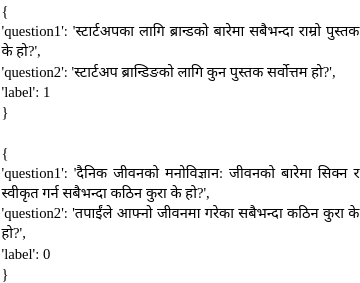}
  \caption{QQP Positive (1) and Negative (0) Sample}
  \label{fig:QQP_exp}
\end{figure}

We incorporated QQP into our benchmark to evaluate paraphrase detection capabilities, which are fundamental for robust language understanding systems. This task is particularly challenging in Nepali due to its morphological richness and limited resources. The dataset allows us to examine whether models can identify semantic equivalence beyond surface-level token matching or basic lexical similarity.

\subsection{Microsoft Research Paraphrase Corpus (MRPC)}
The MRPC dataset consists of pairs of sentences extracted from news sources, labeled as either paraphrase (semantically equivalent) or not. It is widely used to evaluate a model's ability to detect semantic equivalence between sentence pairs, particularly in formal and factual text domains.

MRPC challenges models to identify nuanced semantic similarities beyond superficial word overlap, requiring a deep understanding of sentence structure and meaning in Nepali. Its inclusion in the Nepali benchmark ensures robust evaluation of models’ ability to handle formal news text.

\begin{figure}[H]
  \centering
  \includegraphics[width=1\columnwidth]{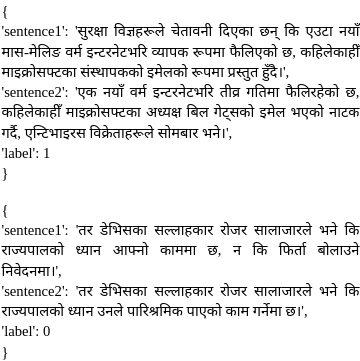}
  \caption{MRPC Positive (1) and Negative (0) Sample}
  \label{fig:MRPC_exp}
\end{figure}

\subsection{Semantic Textual Similarity Benchmark (STS-B)}
The STS-B dataset consists of pairs of sentences annotated with a similarity score that reflects their semantic closeness on a continuous scale, typically from 0 (completely dissimilar) to 5 (semantically equivalent). 

\begin{figure}[H]
  \centering
  \includegraphics[width=1\columnwidth]{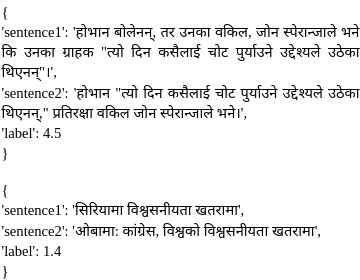}
  \caption{STS-B High (4.5) and Low (1.4) Similarity Sample}
  \label{fig:STSB_exp}
\end{figure}

Unlike binary paraphrase detection tasks, the Semantic Textual Similarity Benchmark (STS-B) requires models to assess fine-grained semantic similarity between sentence pairs on a continuous scale. This makes STS-B a valuable benchmark for evaluating nuanced language understanding. This task challenges models to understand subtle differences and degrees of meaning overlap, which is essential for many real-world applications such as information retrieval, question answering, and summarization.

\subsection{Query-Ad Matching (QADSM)}
The QADSM dataset is incorporated into the NLUE benchmark to assess models’ ability to align semantic meaning between queries and advertisements in a binary classification task.

\begin{figure}[H]
  \centering
  \includegraphics[width=1\columnwidth]{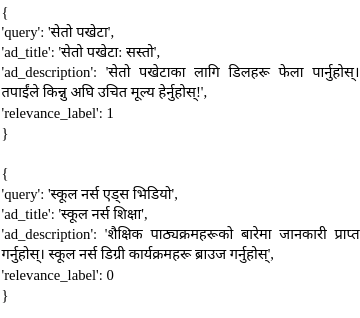}
  \caption{QADSM Positive (1) and Negative (0) Sample}
  \label{fig:QADSM_exp}
\end{figure}

QADSM challenges models to discern semantic relevance beyond superficial keyword matching, requiring a detailed understanding of user intent and contextual meaning in Nepali. This is a critical capability for many real-world applications such as targeted advertising, search result optimization, and personalized content delivery in Nepali, where accurate query-ad alignment is critical.

\subsection{Multi-Genre Natural Language Inference (MNLI)}
The MNLI dataset consists of sentence pairs, each containing a premise and a hypothesis, labeled with one of three classes: entailment, contradiction, or neutral. 

We include MNLI in our benchmark to evaluate a model’s ability to reason about the relationship between sentences across diverse domains. This task extends beyond surface-level similarity, requiring models to capture subtle semantic distinctions, such as entailment, contradiction, and neutrality, which are essential for applications like question answering, summarization, and dialogue systems.

\begin{figure}[H]
  \centering
  \includegraphics[width=1\columnwidth]{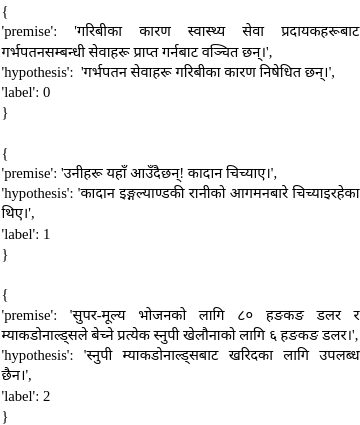}
  \caption{MNLI Entailment (0), Neutral (1) and Contradiction (2) Sample}
  \label{fig:MNLI_exp}
\end{figure}

\subsection{Question-Answering Natural Language Inference (QNLI)}
The QNLI dataset is a sentence pair classification benchmark designed to evaluate a model’s ability to perform natural language inference in the context of question answering. 

\begin{figure}[H]
  \centering
  \includegraphics[width=1\columnwidth]{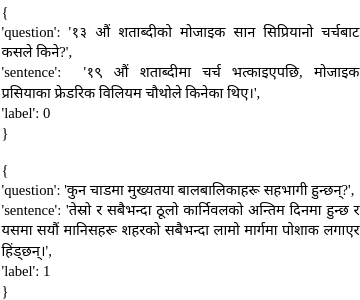}
  \caption{QNLI Entailment (0) and Non-Entailment (1) Sample}
  \label{fig:QNLI_exp}
\end{figure}

We include the QNLI dataset in our benchmark to evaluate a model’s ability to reason over question–answer pairs. This task requires understanding the intent behind a question and determining whether a candidate sentence contains information that answers it, thereby testing the model’s grasp of both question semantics and contextual relevance.

\subsection{Recognizing Textual Entailment (RTE)}
The Recognizing Textual Entailment (RTE) dataset is a benchmark designed to evaluate a model’s ability to determine whether the meaning of one text fragment (the hypothesis) can be inferred from another text fragment (the text). 

\begin{figure}[H]
  \centering
  \includegraphics[width=1\columnwidth]{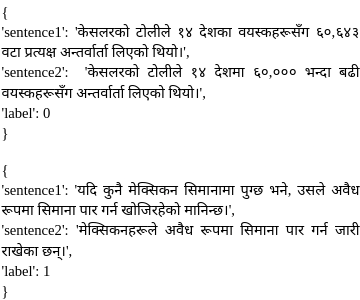}
  \caption{RTE Entailment (0) and Non-Entailment (1) Sample}
  \label{fig:RTE_exp}
\end{figure}

Including RTE in a Nepali benchmark is important because entailment recognition is a core aspect of natural language understanding, especially in low-resource settings where explicit reasoning and semantic alignment are critical. It helps assess whether models trained on Nepali data can capture subtle logical relationships.

\subsection{Co-reference Resolution (CR)}
Co-reference resolution is the task of identifying when two or more expressions in a text refer to the same entity. This is essential for understanding the meaning of a passage, as natural language often relies on pronouns and noun phrases that depend on previous context.

We include the co-reference resolution dataset in our evaluation benchmark to assess a model’s ability to understand and maintain coherence across sentences. In the context of the Nepali language, this task is particularly challenging due to the flexible and context-sensitive nature of referential expressions shaped by discourse. Evaluating models on this task allows us to probe their understanding of entity continuity, pronoun grounding, and broader contextual reasoning.

\begin{figure}[H]
  \centering
  \includegraphics[width=1\columnwidth]{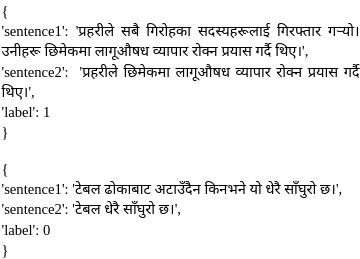}
  \caption{CR Positive (1) and Negative (0) Sample}
  \label{fig:CR_exp}
\end{figure}

\subsection{General Masked Evaluation Task (GMET)}
We developed the General Masked Evaluation Task (GMET) dataset, and it is designed to test whether a model understands context. As the task is to predict masked tokens, we test our models on this task without fine-tuning. Given a mask, any word or phrase could plausibly fit the blank depending on the context, so the model must deeply understand the meaning and structure of the sentence to make an accurate prediction. Including GMET in the benchmark is important because it evaluates general language modeling capabilities, such as contextual comprehension, lexical choice, and syntactic fluency skills that are essential for strong performance across a wide range of downstream tasks in Nepali.

\begin{figure}[H]
  \centering
  \includegraphics[width=1\columnwidth]{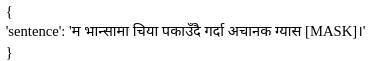}
  \caption{GMET Sample}
  \label{fig:GMET_exp}
\end{figure}

\section{Dataset Translation Approach}
\label{sec:appendixapoint2}
Given the unfunded nature of this research, we relied on personal resources and utilized the APIs of two large language models, GPT-4o-mini and Gemini2.5-flash, to translate datasets into Nepali\footnote{\href{https://github.com/iriis-research/translation-pipeline}{Translation Repository}}. We processed data in batches of 50 to 100 rows, each containing text and its corresponding label, using automated scripts to manage batching, API interactions, and output collection. For tasks requiring nuanced understanding, such as Co-reference Resolution, manual translation and review were employed to ensure accuracy.

\subsection{Translation Problems}
During the translation process, we encountered several challenges:

\subsubsection{Label Mismatch}
\begin{figure}[H]
  \centering
  \includegraphics[width=1\columnwidth]{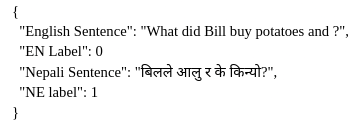}
  \caption{Incomplete coordination in English becomes a well-formed question in Nepali, causing label mismatch.}
  \label{fig:label_msimatch_1}
\end{figure}

\begin{figure}[H]
  \centering
  \includegraphics[width=1\columnwidth]{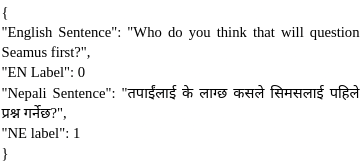}
  \caption{English ungrammaticality from complementizer-trace is absent in Nepali, leading to label mismatch.}
  \label{fig:label_msimatch_2}
\end{figure}

\begin{figure}[H]
  \centering
  \includegraphics[width=1\columnwidth]{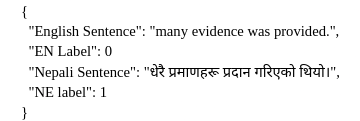}
  \caption{Plural-subject agreement error in English is resolved in Nepali, resulting in label mismatch.}
  \label{fig:label_msimatch_3}
\end{figure}

Despite instructions to preserve labels, models produced correct Nepali translations that differed in meaning from the English source, resulting in label mismatches. In some cases, models also corrected errors in the original English texts, requiring manual review to ensure consistency. Examples are provided in \autoref{fig:label_msimatch_1}, \autoref{fig:label_msimatch_2} and \autoref{fig:label_msimatch_3}. 

\subsubsection{Literal Translations}
\begin{figure}[H]
  \centering
  \includegraphics[width=1\columnwidth]{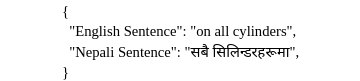}
  \caption{English idiom “on all cylinders” meaning “working perfectly or at full capacity” becomes too literal in Nepali translation, losing its sentiment.}
  \label{fig:literal_1}
\end{figure}

\begin{figure}[H]
  \centering
  \includegraphics[width=1\columnwidth]{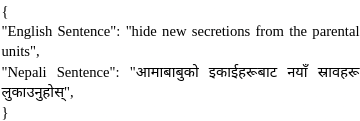}
  \caption{Teen slang that means hiding new secrets from parents, when translated, talks about bodily fluid, losing the context.}
  \label{fig:literal_2}
\end{figure}

\begin{figure}[H]
  \centering
  \includegraphics[width=1\columnwidth]{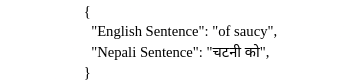}
  \caption{Wordplay lost in translation, English slang becomes nonsensical in Nepali.}
  \label{fig:literal_3}
\end{figure}

\begin{figure}[H]
  \centering
  \includegraphics[width=1\columnwidth]{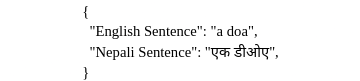}
  \caption{English slang phrase transliterated in Nepali, losing its negative sentiment.}
  \label{fig:literal_4}
\end{figure}

Some translations were overly literal, failing to capture contextual nuances. This issue arises when translation models prioritize word-by-word equivalence rather than interpreting the sentence as a whole. As a result, idiomatic expressions, culturally specific phrases, or context-dependent meanings are mistranslated, leading to loss of intended meaning. Examples are provided in \autoref{fig:literal_1}, \autoref{fig:literal_2}, \autoref{fig:literal_3} and \autoref{fig:literal_4}. 

\subsubsection{Temporal Expression Mismatch}
\begin{figure}[H]
  \centering
  \includegraphics[width=1\columnwidth]{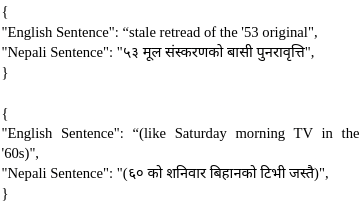}
  \caption{Examples of temporal expression mismatches due to literal translation of abbreviated years.}
  \label{fig:year_express_exp}
\end{figure}

Abbreviated years in English, when translated literally into Nepali numerals, are often misinterpreted as regular numbers rather than references to specific years. This results in a loss of temporal context, which is especially problematic in historical or review texts where accurate time representation is crucial. Such misinterpretations can alter the meaning of the text and reduce the effectiveness of models trained on this data. Careful handling of these expressions is necessary to preserve the intended temporal information in Nepali translations. Examples are provided in \autoref{fig:year_express_exp}.

\subsubsection{Named Entities and Cultural References}

\begin{figure}[H]
  \centering
  \includegraphics[width=1\columnwidth]{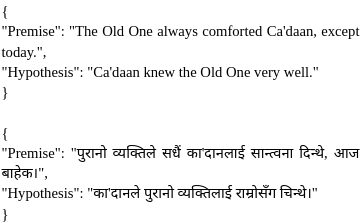}
  \caption{Example of cultural reference mismatch: mythological connotation of "The Old One" is weakened in Nepali translation.}
  \label{fig:cultural_ref_1}
\end{figure}

\begin{figure}[H]
  \centering
  \includegraphics[width=1\columnwidth]{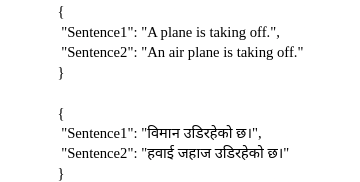}
  \caption{Example of lexical ambiguity in named entities: English terms like "plane" and "airplane" in Nepali have a subtle semantic distinction.}
  \label{fig:cultural_ref_2}
\end{figure}

\begin{figure}[H]
  \centering
  \includegraphics[width=1\columnwidth]{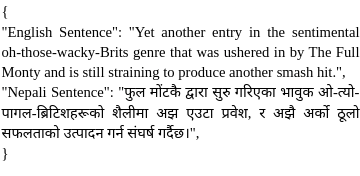}
  \caption{Example of idiomatic mismatch: “oh-those-wacky-Brits” referring to eccentric British cultural traits becomes “oh-those-crazy-British-people,” sounding awkward or negative in Nepali.}
  \label{fig:cultural_ref_3}
\end{figure}

Inconsistent translations of named entities and cultural references often disrupted the semantic integrity of the text and required manual corrections to maintain relevance and consistency within the Nepali context. These inconsistencies, if left uncorrected, could mislead models during training or evaluation. Examples are provided in \autoref{fig:cultural_ref_1}, \autoref{fig:cultural_ref_2} and \autoref{fig:cultural_ref_3}.

\subsection{Suggestions on Translation}
To improve translation quality and accuracy, we recommend the following strategies:

\begin{itemize}
    \item \textbf{Use detailed prompts:} Instruct models to translate into clear, natural Nepali while preserving the original meaning and sentence structure.
    
    \item \textbf{Handle untranslatable terms:} For words or phrases without direct Nepali equivalents, allow romanization as a fallback strategy.
    
    \item \textbf{Batch size optimization:} Process \textbf{50--100 examples per API request} (assuming each has 80-100 tokens) to balance efficiency with translation quality. Avoid exceeding 100 examples to prevent degradation.
    
    \item \textbf{Class-wise translation:} Translate examples belonging to the same class in separate requests, and assign labels \textbf{after} translation to avoid mismatches due to grammatical differences across languages.
    
    \item \textbf{Ensure output consistency:} Implement automated checks to verify that the \textbf{number of translated outputs matches the input examples}, minimizing the risk of data loss during batch processing.
\end{itemize}

\section{Dataset Quality}
\label{sec:appendixapoint3}

\subsection{Multilingual Content Filtering}
We employed automated language detection to identify and remove any English or non-Nepali text remnants from the translated outputs. This filtering process ensures the purity of the translation by flagging code-switching instances, incomplete translations, or processing errors that leave the artifacts of the source language.

\subsection{Statistical Quality Sampling with Manual Validation}
A random sampling approach was used to select 1\% of the translated corpus for manual quality assessment by native Nepali speakers. Each sampled translation was evaluated using standardized rubrics covering adequacy, fluency, and cultural appropriateness. It was decided that if more than 10\% of the samples were found to be unacceptable, retranslation would be performed with an updated prompt. However, no such cases were encountered, indicating high confidence in the translation quality achieved using GPT-4o-mini and Gemini-2.5-flash for Nepali language translation.

\subsection{Bidirectional Translation Validation (Back-translation)}
Back-translation validation was performed on a randomly sampled 1\% subset by translating Nepali outputs back to English using a different translation system. Semantic preservation was measured through automated similarity metrics, including BLEU scores between original and back-translated English texts. Some divergences indicated potential quality issues such as semantic drift or ambiguity resolution errors in the forward translations, but no significant concerns were seen.

\section{Experiment Stats}
\label{sec:appendixapoint4}

We utilized ~1,200 GPU hours on NVIDIA T4 GPUs for our experiments. This includes fine-tuning 10 distinct model variants on 11 benchmark datasets, on the configurations outlined in the experiments (\autoref{sec:experiments}).

\section{Hyperparameter Settings}
\label{sec:appendixapoint5}

This section details the optimal hyperparameters identified for each model and dataset combination in our benchmark evaluation for reproducibility. Tuning config is written in the following order: Model Config, Learning Rate, Epoch, Batch Size.

For model config, see \autoref{fig:training_config_with_FC} and \autoref{fig:training_config_without_FC}.

\subsection{Single Sentence Tasks}
Best hyperparameter settings (model config, learning rate, epochs, batch size) for each model on SA (Sentiment Analysis), CoLA (Corpus of Linguistic Acceptability), and WG (WinoGrande) tasks are reported in \autoref{tab:hyperparams-sa-cola-wg}.

\begin{table*}[htbp]
\centering
\resizebox{\textwidth}{!}{%
\begin{tabular}{lcccc}
\hline
\textbf{Model} & \textbf{Params} 
& \textbf{SA} 
& \textbf{CoLA} 
& \textbf{WG} \\
\hline
Distilbert-Nepali~\cite{maskey-etal-2022-nepali}   & 67M  & I, 2e-5, 3, 16 & I, 2e-5, 5, 16 & II, 2e-5, 8, 32 \\
NepBERT~\cite{rajan2021github}                    & 82M  & I, 2e-5, 2, 16 & I, 2e-5, 8, 16 & II, 2e-5, 4, 32 \\
NepaliBERT~\cite{Pudasaini2023}                   & 110M & I, 2e-5, 2, 16 & I, 2e-5, 8, 16 & II, 2e-5, 5, 32 \\
BERT Nepali~\cite{thapa-etal-2025-development}     & 110M & I, 2e-5, 2, 16 & I, 2e-5, 10, 16 & II, 2e-5, 10, 32 \\
NepBERTa~\cite{timilsina-etal-2022-nepberta}       & 110M & I, 2e-5, 2, 16 & I, 2e-5, 8, 16 & I, 2e-5, 9, 32 \\
RoBERTa Nepali~\cite{thapa-etal-2025-development}  & 125M & I, 2e-5, 2, 16 & I, 2e-5, 9, 16 & I, 2e-5, 9, 32 \\
DeBERTa-Nepali~\cite{maskey-etal-2022-nepali}      & 139M & I, 2e-5, 3, 16 & I, 2e-5, 5, 16 & II, 2e-5, 8, 32 \\
Multilingual BERT~\cite{Devlin2019}               & 172M & I, 2e-5, 3, 16 & I, 2e-5, 10, 16 & II, 2e-5, 10, 32 \\
XLM-R Base~\cite{conneau-etal-2020-unsupervised}   & 270M & I, 2e-5, 3, 16 & I, 2e-5, 5, 16 & I, 2e-5, 8, 32 \\
m-DeBERTa-v3~\cite{he2023debertav3improvingdebertausing} & 276M & I, 2e-5, 3, 16 & I, 2e-5, 5, 16 & I, 2e-5, 8, 32 \\
\hline
\end{tabular}
}
\caption{\label{tab:hyperparams-sa-cola-wg}
Best hyperparameter settings for Single Sentence Tasks.
}
\end{table*}

\begin{table*}[htbp]
\centering
\resizebox{\textwidth}{!}{%
\begin{tabular}{lccccc}
\hline
\textbf{Model} & \textbf{Params} 
& \textbf{QQP} 
& \textbf{MRPC} 
& \textbf{STS-B} 
& \textbf{QADSM} \\
\hline
Distilbert-Nepali~\cite{maskey-etal-2022-nepali}   & 67M  & VI, 2e-5, 3, 16  & VI, 2e-5, 4, 16  & I, 2e-5, 15, 16 & VI, 2e-5, 4, 16 \\
NepBERT~\cite{rajan2021github}                    & 82M  & VI, 2e-5, 2, 16  & VI, 2e-5, 3, 16  & I, 2e-5, 15, 16 & VI, 2e-5, 3, 16 \\
NepaliBERT~\cite{Pudasaini2023}                   & 110M & VI, 2e-5, 4, 16  & VII, 2e-5, 5, 16 & I, 2e-5, 14, 16 & VI, 2e-5, 5, 16 \\
BERT Nepali~\cite{thapa-etal-2025-development}     & 110M & VI, 2e-5, 2, 16  & VII, 2e-5, 5, 16 & I, 2e-5, 15, 16 & VI, 2e-5, 5, 16 \\
NepBERTa~\cite{timilsina-etal-2022-nepberta}       & 110M & VI, 2e-5, 4, 16  & VII, 2e-5, 4, 16 & I, 2e-5, 15, 16 & VI, 2e-5, 3, 32 \\
RoBERTa Nepali~\cite{thapa-etal-2025-development}  & 125M & VI, 2e-5, 2, 16  & VI, 2e-5, 6, 16  & II, 2e-5, 12, 8 & VI, 2e-5, 3, 32 \\
DeBERTa-Nepali~\cite{maskey-etal-2022-nepali}      & 139M & VI, 2e-5, 2, 16  & VII, 2e-5, 4, 16 & I, 2e-5, 5, 16  & VI, 2e-5, 5, 16 \\
Multilingual BERT~\cite{Devlin2019}               & 172M & VI, 2e-5, 2, 16  & VII, 2e-5, 3, 16 & I, 2e-5, 5, 16  & VI, 2e-5, 5, 16 \\
XLM-R Base~\cite{conneau-etal-2020-unsupervised}   & 270M & VI, 2e-5, 2, 16  & VI, 2e-5, 4, 16  & I, 2e-5, 13, 16 & VI, 2e-5, 3, 32 \\
m-DeBERTa-v3~\cite{he2023debertav3improvingdebertausing} & 276M & VI, 2e-5, 3, 16  & VII, 2e-5, 6, 16 & I, 2e-5, 14, 16 & VI, 2e-5, 3, 16 \\
\hline
\end{tabular}
}
\caption{\label{tab:hyperparams-qqp-mrpc-stsb-qadsm}
Best hyperparameter settings for Similarity and Paraphrase Tasks.
}
\end{table*}

\begin{table*}[htbp]
\centering
\resizebox{\textwidth}{!}{%
\begin{tabular}{lccccc}
\hline
\textbf{Model} & \textbf{Params} 
& \textbf{MNLI} 
& \textbf{QNLI} 
& \textbf{RTE} 
& \textbf{CR} \\
\hline
Distilbert-Nepali~\cite{maskey-etal-2022-nepali} & 67M & VI, 2e-5, 3, 16 & VI, 2e-5, 2, 16 & V, 2e-5, 10, 32 & V, 1e-5, 4, 16 \\
NepBERT~\cite{rajan2021github} & 82M & VI, 2e-5, 4, 16 & VI, 2e-5, 3, 16 & V, 2e-5, 12, 32 & V, 1e-5, 3, 16 \\
NepaliBERT~\cite{Pudasaini2023} & 110M & VI, 2e-5, 3, 16 & VI, 2e-5, 2, 16 & V, 2e-5, 9, 32 & IV, 1e-5, 3, 16 \\
BERT Nepali~\cite{thapa-etal-2025-development} & 110M & VI, 2e-5, 3, 16 & VI, 2e-5, 2, 16 & V, 2e-5, 11, 32 & V, 1e-5, 2, 16 \\
NepBERTa~\cite{timilsina-etal-2022-nepberta} & 110M & VI, 2e-5, 3, 16 & VI, 2e-5, 2, 16 & V, 2e-5, 10, 32 & V, 1e-5, 2, 32 \\
RoBERTa Nepali~\cite{thapa-etal-2025-development} & 125M & VII, 2e-5, 7, 16 & VI, 2e-5, 2, 16 & V, 2e-5, 15, 32 & V, 2e-5, 5, 32 \\
DeBERTa-Nepali~\cite{maskey-etal-2022-nepali}  & 139M & VII, 2e-5, 5, 16 & VI, 2e-5, 4, 16 & V, 2e-5, 10, 32 & IV, 2e-5, 5, 32 \\
Multilingual BERT~\cite{Devlin2019} & 172M & VI, 2e-5, 3, 16 & VI, 2e-5, 2, 16 & V, 2e-5, 10, 32 & IV, 2e-5, 3, 32 \\
XLM-R Base~\cite{conneau-etal-2020-unsupervised} & 270M & VII, 2e-5, 5, 16 & VI, 2e-5, 3, 16 & V, 2e-5, 15, 32 & IV, 2e-5, 3, 32 \\
m-DeBERTa-v3~\cite{he2023debertav3improvingdebertausing} & 276M & VI, 2e-5, 3, 16 & VI, 2e-5, 2, 16 & IV, 2e-5, 12, 32 & IV, 1e-5, 4, 8 \\
\hline
\end{tabular}
}
\caption{\label{tab:hyperparams-mnli-qnli-rte-cr}
Best hyperparameter settings for Inference Tasks.
}
\end{table*}

\subsection{Similarity and Paraphrase Tasks}
Best hyperparameter settings (model config, learning rate, epochs, batch size) for each model on QQP (Quora Question Pairs), MRPC (Microsoft Research Paraphrase Corpus), STS-B (Semantic Textual Similarity Benchmark), and QADSM (Query Ad Matching) tasks are reported in \autoref{tab:hyperparams-qqp-mrpc-stsb-qadsm}.

\subsection{Inference Tasks}
Best hyperparameter settings (model config, learning rate, epochs, batch size) for each model on MNLI (Multi-Genre Natural Language Inference), QNLI (Question-answering Natural Language Inference), RTE (Recognizing Textual Entailment), and CR (Co-reference Resolution) tasks are reported in \autoref{tab:hyperparams-mnli-qnli-rte-cr}.

\section{More on GMET}
\label{sec:appendixapoint6}

\subsection{Dataset Categories}
The GMET dataset is organized into the following 75 categories, grouped into seven thematic areas, presented here in English:

\begin{itemize}
  \item \textbf{Daily Life \& Home:}  
  Family, House, Kitchen, Food, Clothing, Market, Shop, Daily Routine, Furniture, Health, Dream, Cleanliness, Medicine

  \item \textbf{Nature \& Environment:}  
  Weather, Animals, Birds, Insects, Fruits, Vegetables, Trees, Flowers, Nature, Water, River, Mountain, Forest, Sky, Earth, Ocean/Sea, Weather Conditions

  \item \textbf{Society \& Culture:}  
  School, Village, City, Sports, Festivals, Music, Art, Friendship, Society, Language, Nepali Culture, Movies, Books

  \item \textbf{Concepts \& Knowledge:}  
  Colors, Body Parts, Time, Numbers, Days of the Week, Months, Feelings, Shapes, Directions, Senses, Opposites, Geography, Science

  \item \textbf{Work \& Activities:}  
  Work, Professions, Agriculture, Learning, Cooking, Hobbies, Communication, Travel

  \item \textbf{Broader World:}  
  Transportation, Money, History, Government, Technology, Space (sun, moon), Tools, Materials (wood, metal)

  \item \textbf{Nepal Specific:}  
  Geography of Nepal, Animals of Nepal, Festivals of Nepal
\end{itemize}

\subsection{Evaluation Metrics for GMET}
The General Masked Evaluation Task (GMET) employs two primary metrics to assess language model performance: overall accuracy and a combined score. These metrics are formalized as follows: 

\subsubsection{Overall Accuracy}
The overall accuracy is defined as the proportion of correct predictions across all sentences in the dataset. Given a dataset with N = 1500 sentences, where each prediction is scored as 1 (correct) or 0 (incorrect), let $s_{i}$ represent the score for the $i^{th}$ sentence. The overall accuracy A
is calculated as:
$$
A = \frac{1}{N} \sum_{i=1}^{N} s_i
$$

\subsubsection{Combined Accuracy}
The combined score integrates the overall accuracy with an equality score that measures consistency across categories. The dataset is divided into K = 75 categories, with each category containing 20 sentences. For the $k^{th}$ category, the category-wise accuracy $A_k$ is computed as: 

$$
A_k = \frac{1}{20} \sum_{i=1}^{20} s_{i, k}
$$

where $s_{i,k}$ is the score for the $i^{th}$ sentence in the $k^{th}$ category. 
The standard deviation of the category-wise accuracies, $\sigma$, is calculated as: 

$$
\sigma = \sqrt{\frac{1}{K} \sum_{i=1}^K (A_k - \bar{A})^2}
$$

where $\bar{A}$ is the mean of the category-wise accuracies:
$$
\bar{A} = \frac{1}{K} \sum_{i=1}^{K} A_k
$$

The equality score, $E$, is derived by transforming the standard deviation to map lower deviation values to higher scores, with the score ranging between 0 and 1. The equality score is defined as:

$$
E = e^{-\sigma}
$$

This function ensures that lower standard deviations (indicating more consistent performance across categories) yield higher equality scores. In the edge case of a single category, where $\sigma$ is undefined, $E$ is set to 1. The combined score, $C$, is then computed as the product of the overall accuracy and the equality score:

$$
C = A.E
$$

This combined score balances overall performance with consistency, penalizing models that exhibit uneven performance across the diverse linguistic and topical categories of the GMET dataset.

\section{Performance Visualization of Individual Models on each task}
\label{sec:appendixapoint7}

\begin{figure}[H]
  \centering
  \includegraphics[width=0.85\columnwidth]{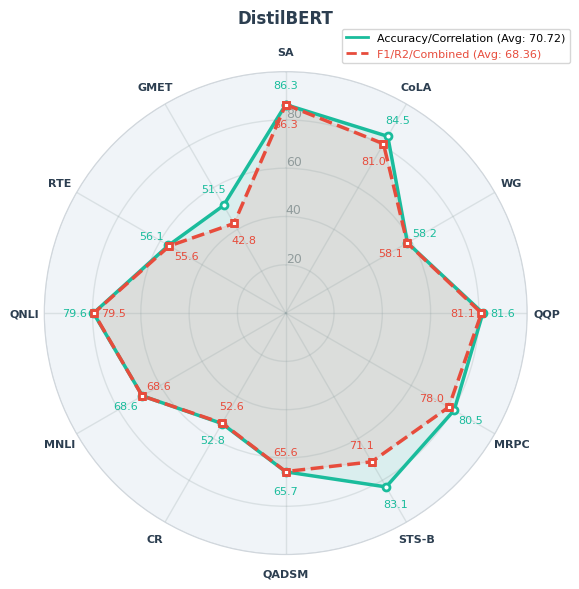}
  \caption{DistillBERT across all tasks}
  \label{fig:distillbert}
\end{figure}

\begin{figure}[H]
  \centering
  \includegraphics[width=0.85\columnwidth]{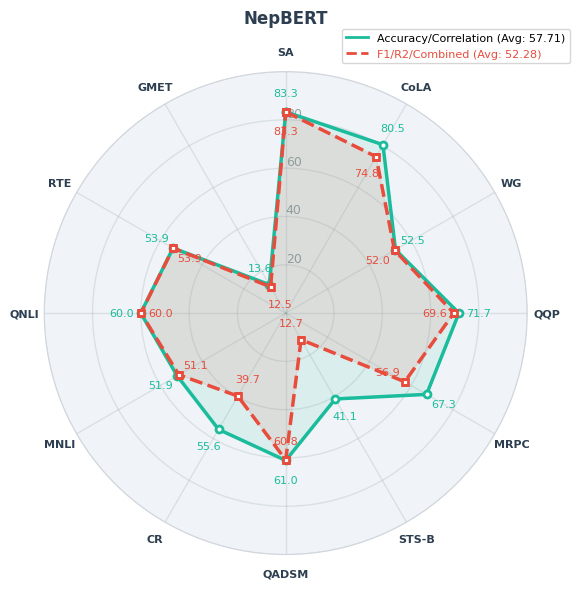}
  \caption{NepBERT across all tasks}
  \label{fig:nepbert}
\end{figure}

\begin{figure}[H]
  \centering
  \includegraphics[width=0.85\columnwidth]{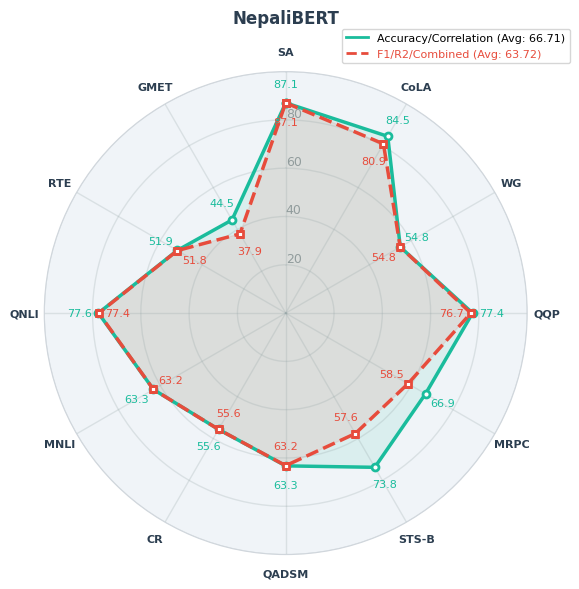}
  \caption{NepaliBERT across all tasks}
  \label{fig:nepalibert}
\end{figure}

\begin{figure}[H]
  \centering
  \includegraphics[width=0.85\columnwidth]{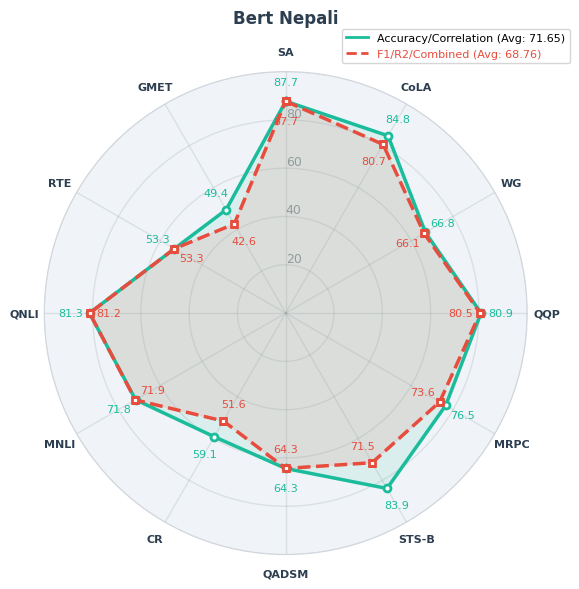}
  \caption{BERT Nepali across all tasks}
  \label{fig:bertnepali}
\end{figure}

\begin{figure}[H]
  \centering
  \includegraphics[width=0.85\columnwidth]{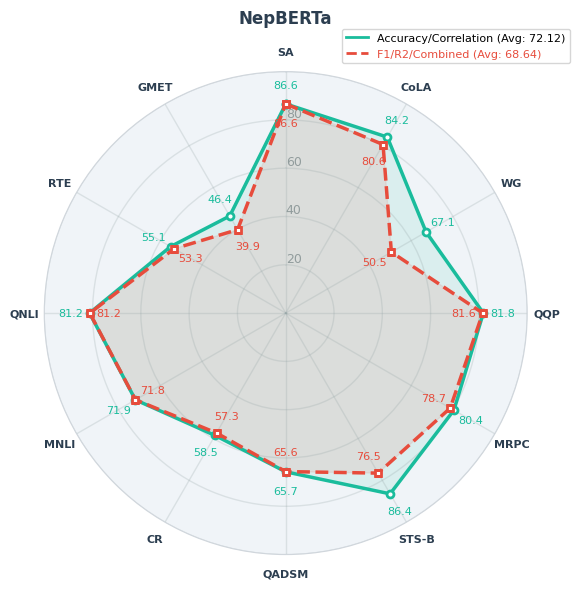}
  \caption{NepBERTa acrosss all tasks}
  \label{fig:nepberta}
\end{figure}

\begin{figure}[H]
  \centering
  \includegraphics[width=0.85\columnwidth]{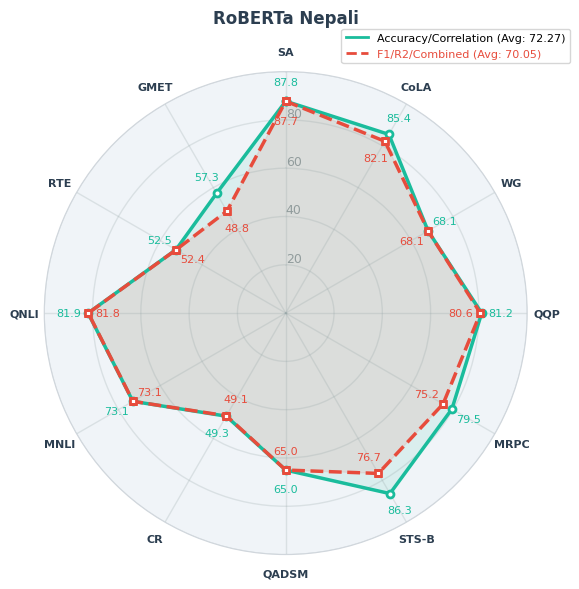}
  \caption{RoBERTa Nepali across all tasks}
  \label{fig:robertanepali}
\end{figure}

\begin{figure}[H]
  \centering
  \includegraphics[width=0.85\columnwidth]{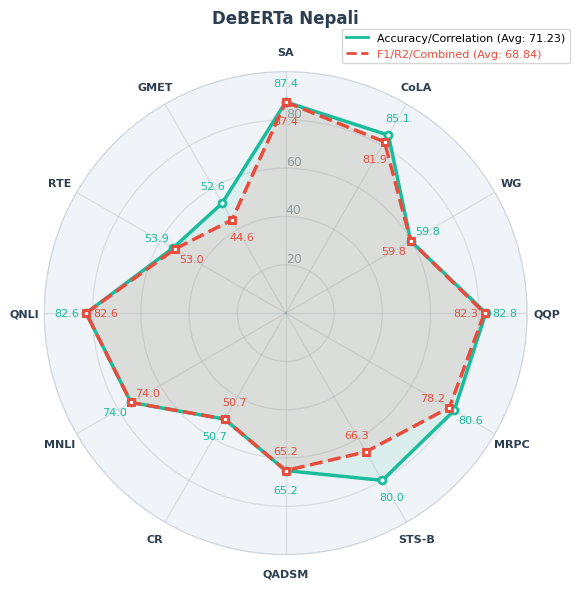}
  \caption{DeBERTa Nepali across all tasks}
  \label{fig:debertanepali}
\end{figure}

\begin{figure}[H]
  \centering
  \includegraphics[width=0.85\columnwidth]{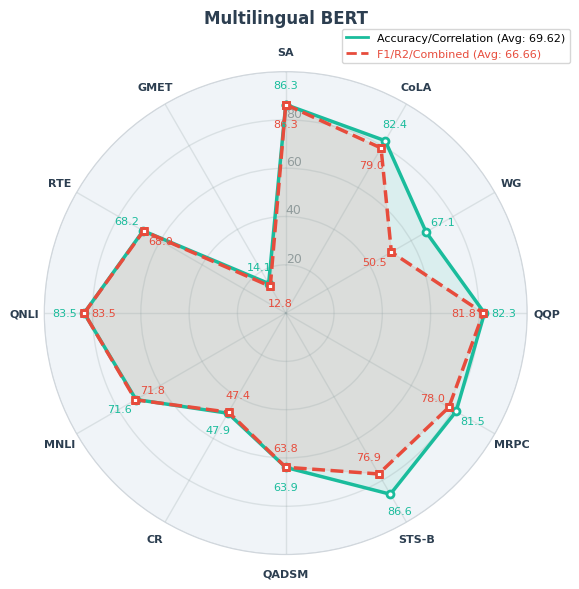}
  \caption{Multilingual BERT across all tasks}
  \label{fig:multilingualbert}
\end{figure}

\begin{figure}[H]
  \centering
  \includegraphics[width=0.85\columnwidth]{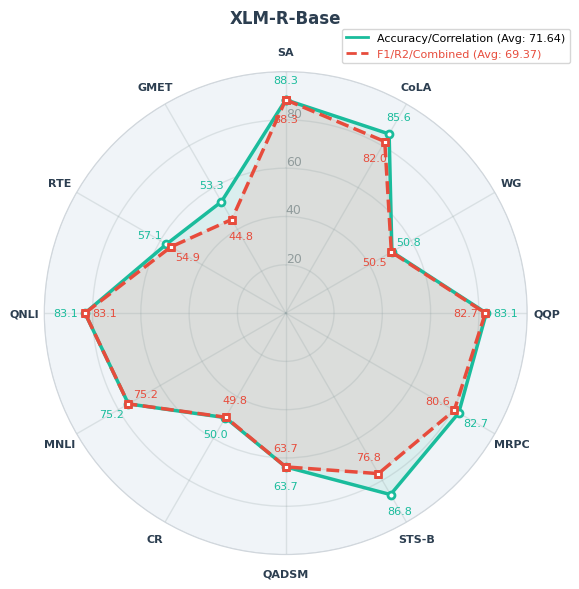}
  \caption{XLM-R-Base across all tasks}
  \label{fig:xlm}
\end{figure}

\begin{figure}[H]
  \centering
  \includegraphics[width=0.85\columnwidth]{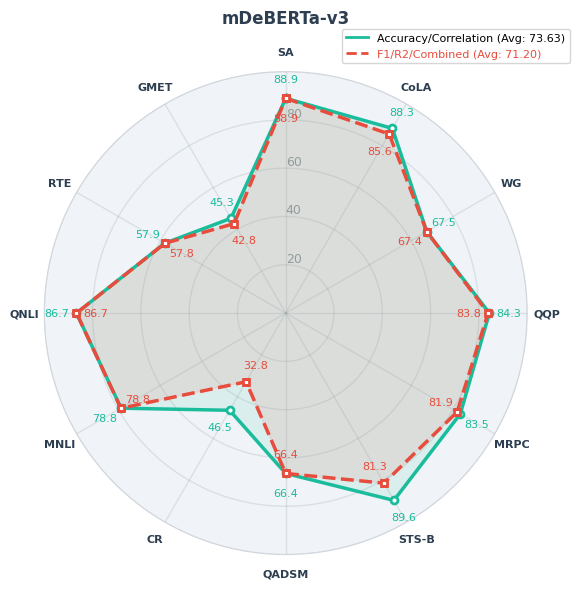}
  \caption{mDeBERTa-v3 across all tasks}
  \label{fig:mdebertav3}
\end{figure}

\end{document}